\documentclass[10pt,twocolumn,letterpaper]{article}

\usepackage{cvpr}
\usepackage{times}
\usepackage{epsfig}
\usepackage{graphicx}
\usepackage{mathtools}
\usepackage{amsmath}
\usepackage{amssymb}
\usepackage{multirow}
\usepackage{algorithm}
\usepackage{algorithmic}

\DeclareMathOperator*{\argminB}{argmin}   

\usepackage[breaklinks=true,bookmarks=false]{hyperref}

\cvprfinalcopy 

\def\cvprPaperID{****} 

\begin{document}

\title{SYQ: Learning Symmetric Quantization For Efficient Deep Neural Networks}

\author{Julian Faraone* \quad Nicholas Fraser$^{\#}$ \quad Michaela Blott$^{\#}$ \quad Philip H.W. Leong*\\
The University of Sydney*\\
Xilinx Research Labs$^{\#}$\\
{\tt\small (julian.faraone, philip.leong)@sydney.edu.au} \quad \tt\small(nfraser, mblott)@xilinx.com
}

\maketitle

\begin{abstract}
Inference for state-of-the-art deep neural networks is computationally expensive, making them difficult to deploy on constrained hardware environments. An efficient way to reduce this complexity is to quantize the weight parameters and/or activations during training by approximating their distributions with a limited entry codebook. For very low-precisions, such as binary or ternary networks with 1-8-bit activations, the information loss from quantization leads to significant accuracy degradation due to large gradient mismatches between the forward and backward functions. In this paper, we introduce a quantization method to reduce this loss by learning a symmetric codebook for particular weight subgroups. These subgroups are determined based on their locality in the weight matrix, such that the hardware simplicity of the low-precision representations is preserved. Empirically, we show that symmetric quantization can substantially improve accuracy for networks with extremely low-precision weights and activations. We also demonstrate that this representation imposes minimal or no hardware implications to more coarse-grained approaches. Source code is available at https://www.github.com/julianfaraone/SYQ.
\end{abstract}

\section{Introduction}
Deep Neural Networks (DNNs) have produced state-of-the-art results in applications such as computer vision \cite{Krizhevsky:2012:ICD:2999134.2999257}, natural language processing \cite{DBLP:journals/corr/abs-1103-0398} and object detection \cite{Viola01robustreal-time}. As their size continues to grow to improve prediction capabilities, their memory and computational requirements also scales, making them increasingly difficult to deploy on embedded systems. For example, \cite{Krizhevsky:2012:ICD:2999134.2999257} achieved state-of-art-results on the ImageNet challenge using AlexNet which required 240MB of storage and 1.45 billion operations to compute inference per image. Several methods of compression \cite{DBLP:journals/corr/HanMD15}, quantization \cite{DBLP:journals/corr/ChenWTWC15} and dimensionality reduction \cite{DBLP:journals/corr/abs-1708-00630} have been applied to reduce these demands, with promising results. This demonstrates the over-parametrization and redundancies in DNNs and poses an opportunity for utilizing regularization to make their representations more amenable to hardware implementations. 

In particular, low-precision neural networks reduce both memory and computational requirements whilst achieving accuracies comparable to floating point \cite{Gupta:2015:DLL:3045118.3045303}. For extremely low-precisions, such as binary and/or ternary weight representations and 1-8 bits for activations, most of the multiply-accumulate (MAC) operations can be replaced by simple bitwise operations. This translates to massive reductions in storage requirements and spatial complexity in hardware. Additionally, large power savings and speed gains are achieved when networks can fit in on-chip memory. The issue is that a large reduction in precision, leads to large information loss which incurs significant accuracy degradation, especially for complex datasets such as ImageNet \cite{Russakovsky:2015:ILS:2846547.2846559}. Ideally, we can train networks which have both high prediction capabilities and minimal computational complexity.

DNN training is an iterative process which has a feedforward path to compute the output and a backpropagation path to calculate gradients and update its parameters for learning. Low-precision networks involve having a set of full-precision weights which are quantized before computing inference. As the quantization functions are piecewise and constant, the gradients of quantized weights are calculated and applied to update their corresponding full-precision weights. Similarly, derivatives of quantized activations are calculated by using a non-constant differentiable approximation function. This type of training was first proposed as the Straight Through Estimator (STE) \cite{DBLP:journals/corr/BengioLC13} which suggested the use of a nonzero derivative approximation to functions which are non-differentiable or have zero derivatives everywhere. The problem is that without an accurate estimator for weights and activations, there exists a significant gradient mismatch which impinges on learning. Seemingly, as discussed in \cite{DBLP:journals/corr/MiyashitaLM16}, activations are more robust to quantization than weights for image classification problems due to weight reuse in Convolutional (CONV) layers affecting multiple operations. To overcome this, methods such as increasing the weight codebook by applying a scaling coefficient to all weights in a layer, provides better approximations for weight distributions and greater model capacity \cite{DBLP:journals/corr/LiL16}. This is computationally inexpensive and can be represented as multiplying each weight layer's matrix by a diagonal scalar matrix which only requires storage of one value. Applying fine-grained scaling coefficients has also been shown to improve accuracy by increasing model capacity \cite{DBLP:journals/corr/MellempudiKM0KD17}, \cite{DBLP:journals/corr/RastegariORF16}. The problem with all of these fine-grained approaches is either large storage requirements for the scaling coefficients or high computational complexity due to irregular codebook indices. In this paper we present Learning Symmetric Quantization (SYQ), a method to design binary/ternary networks with fine-grained scaling coefficients which preserve these complexities. We do this by learning a symmetric weight codebook via gradient-based optimizations which enables a minimally-sized square diagonal scalar matrix representation. To reduce the large information loss from CONV layer quantization, we use a more fine-grained pixel/row-wise scaling approach, rather than layer-wise scaling in Fully-Connected (FC) layers. In the process, we significantly close the accuracy gap for low-precision networks to their floating point counterpart, whilst preserving their efficient computational structures. Our work makes the following contributions:

\begin{itemize}
	\item Our approach significantly improves the ability of convolutional weights to learn low-precision representations. This is useful as most layers in modern network architectures consist of convolutions which are typically the least redundant layers.
	\item The proposed method reduces the computational complexity of traditional fine-grained low-precision scaling and imposes minimal hardware costs to layer-wise scaling. 
	\item On state-of-the-art networks such as AlexNet, ResNet and VGG, our method is empirically shown to improve accuracy for 1-2 bit weights and 2-8 bit activations.
\end{itemize}
\section{Related Work}
Most methods for training low-precision DNNs maintain a set of full precision weights that are deterministically or stochastically quantized during forward or backward propagation. Gradient updates computed with the quantized weights are then applied to the full
precision weights \cite{DBLP:journals/corr/CourbariauxBD15},  \cite{DBLP:journals/corr/HubaraCSEB16}, \cite{DBLP:journals/corr/LinCMB15}. To produce state-of-the-art results on larger models,  \cite{DBLP:journals/corr/RastegariORF16} proposed scaling the quantized weights by the expectation of real-valued weights to recover the dynamic range of each layer. \cite{DBLP:journals/corr/LiL16} also implemented a similar technique for ternary networks and optimised a non-zero quantization threshold as a function of the weight expectation. Other gradient-based optimization methods for the scaling coefficient have been introduced \cite{DBLP:journals/corr/ZhuHMD16}. Other methods of quantization have also been implemented, i.e. re-training networks using incremental weight subgrouping to produce no accuracy loss for 5 bit weights \cite{DBLP:journals/corr/ZhouYGXC17}. Multiple binarizations and a scaling layer were described in \cite{Tang2017HowTT} to improve accuracy and binarize the last layer. Logarithmic data representations were used to approximate the non-uniform distribution of the weights, activations and gradients down to 3-bits with negligible accuracy loss~\cite{DBLP:journals/corr/MiyashitaLM16}. Activations quantization has also been investigated with frameworks created for varying activation bitwidths \cite{DBLP:journals/corr/ZhouNZWWZ16} and both weights and activations \cite{Park_2017_CVPR}. Improving the network learnability under low-precision weights and activations was analysed in \cite{DBLP:journals/corr/CaiHSV17}. More fine-grained approaches of quantization have effectively clustered weights or grouped filters together and quantize differently based on their statistical distributions \cite{Duan_2017_CVPR}, \cite{DBLP:journals/corr/MellempudiKM0KD17} . Increasing model capacity by applying scaling coefficients to positive and negative values separately was proposed in \cite{DBLP:journals/corr/ZhuHMD16}. Furthermore, sparse representations were used as regularization to make networks more amenable to hardware \cite{DBLP:journals/corr/abs-1709-06262}. Also, many low-precision DNN hardware implementations have been published \cite{DBLP:journals/corr/VenkateshNM16}, \cite{DBLP:journals/corr/HanLMPPHD16}. For example, FINN \cite{Fraser:2017:SBN:3029580.3029586}, \cite{DBLP:journals/corr/UmurogluFGBLJV16} demonstrated the performance gains of being able to store all network weights in on-chip memory by implementing binarized neural networks on FPGAs.

\section{Low-Precision Networks}
In this section we discuss the motivations behind our work and fundamentals of low-precision neural networks.

\subsection{Motivation}
\begin{figure}
	\vspace{-18pt}
	\includegraphics[width=3.2in]{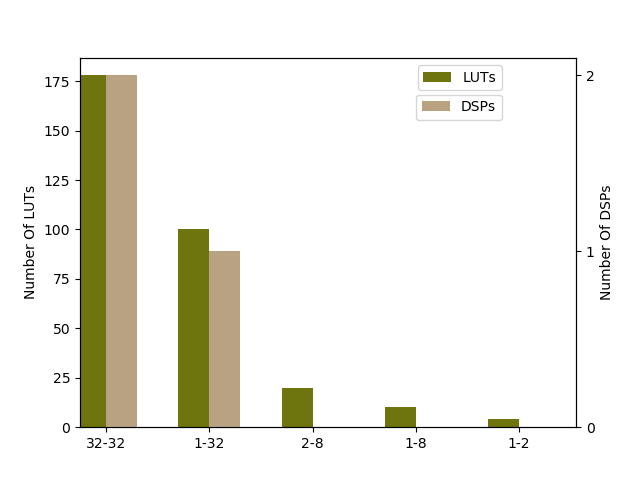}
	\caption{The average cost per MAC operation on an FPGA device for different bitwidths (weight-activation)}
	\label{fig:Motivation}
	\vspace{-15pt}
\end{figure}

Each layer of a DNN computes dot products between weight parameters and its input values. We can represent the output of each hidden unit $h$, as:
\begin{align}
& h= g(\textbf{\textit{w}}^T\textbf{\textit{x}})
\label{eq:DNN_act}
\end{align}
where $g$ is an element-wise nonlinear activation function, $\textbf{\textit{x}}\in\mathbb{R}^{i.w.h}$ is the input vector, and $\textbf{\textit{w}}\in\mathbb{R}^{i.w.h}$ provides the weight vector of a linear transformation. This computation is repeated throughout the network, therefore overall model complexity is dependant on its structure. As modern networks continue to get deeper/wider, model complexity becomes problematic for their applicability on constrained hardware environments. A solution is to efficiently quantize both weights and activations to very low-precisions (1-8 bits) with negligible or no accuracy loss. In doing so, the arithmetic operations are greatly simplified, reducing both computational and resource complexity. In the binary/ternary weight case, MACs are replaced by bit operations. For example, Figure \ref{fig:Motivation} shows average resource usage on Field Programmable Gate Array (FPGA) hardware to implement a MAC operation under different precisions, which scales quadratically with the multiplier size at $\mathcal{O}(k^2)$ where $k$ is the number of bits\footnote {Results are obtained from instantiating MAC modules using Vivado}.   
As shown, no high precision multipliers (known as DSPs on an FPGA) are required for precisions less than or equal to ternary weights and 8-bit activations. Furthermore, the logic element (known as LUTs on an FPGA) requirement reduces proportionally with both weight and activation precisions. Additionally, the storage requirements for both weights and activations is reduced by $8-32\times$. This significantly improves the network's ability to fit in on-chip memory and constrained hardware environments, and broadens the applicability of DNNs. 

For a CONV layer, all weights are typically represented as a tensor $\textbf{W}_l\in\mathbb{R}^{K\times K\times I\times N}$ where $K$ is the filter size, $I$ is the number of input feature maps and $N$, the number of output feature maps. In low-precision networks, each weight layer $l$ can typically be represented by a diagonal scalar matrix  $\mathbf{\boldsymbol{\alpha_l}}$ multiplied by quantized weight matrix $\textbf{\textit{Q}}_l$ and ideally $\textbf{\textit{W}}_l \approx \mathbf{\boldsymbol{\alpha_l}}\textbf{\textit{Q}}_l$. Also, the activation function $g$ can be approximated using a piecewise constant activation function $G$. In our proposed method, we observe that by ensuring quantization levels for $\textbf{\textit{W}}$ are symmetric around zero, we can construct efficient square diagonal matrix representations of $\mathbf{\boldsymbol{\alpha_l}}$, which enable fine-grained quantization whilst having minimal memory requirements (of size $K$ or $K^2$). This translates to a reduction in overall model complexity and high prediction capabilities. Although, we restrict ourselves by structured matrices and low-precision weights and activations, the network efficiently captures information through our gradient-based symmetric quantizer which learns the diagonal elements of  $\mathbf{\boldsymbol{\alpha_l}}$ during training.


\subsection{Weight Quantization}
For low-precision DNNs, the distribution of full precision weight matrices for each layer $\textbf{\textit{W}}_{l}$ are approximated by a function $f$, resulting in a quantized weight matrix $\textbf{\textit{Q}}_l$:
\begin{align}
& \textit{Q}_{l_{i,j}} = f(\textit{W}_l)_{i,j}
\end{align}
for $\textit{W}_{l_{i,j}}\in\mathbb{R}$ and $\textit{Q}_{l_{i,j}}\in\mathbb{C}$. The codebook $\mathbb{C}=\big\{c_1, c_2, ..., c_r\big\}$ is a set of all possible values for $\textit{Q}_{l_{i,j}}$  where $c_i\in\mathbb{R}$ and $i\in\mathbb{R}^+$ represent each codebook value and index respectively. For example, binary and ternary weight spaces have  $\mathbb{C}=\big\{-1, +1\big\}$ and $\mathbb{C}=\big\{-1, 0, +1\big\}$ respectively. Efficient functions for binarizing and ternarizing weight parameters have been proposed as piecewise constant functions in \cite{DBLP:journals/corr/LiL16}, such that:
\begin{align}
& \textbf{\textit{Q}}_{l} = sign(\textbf{\textit{W}}_l) \odot \textbf{\textit{M}}_l
\label{weight_quant}
\end{align}
\vspace{-5pt}
with,
\begin{align}
& \textit{M}_{l_{i,j}} = 
\begin{cases}
& \text{1} \quad \text{ if } \quad \left|\textit{W}_{l_{i,j}}\right| \geq \eta_l\\
& \text{0} \quad \text{ if } \quad -\eta_l < \textit{W}_{l_{i,j}} < \eta_l
\end{cases}
\label{mask}
\end{align}
where $\textbf{\textit{M}}$ represents a masking matrix, $\eta$ is the quantization threshold hyperparameter. $\eta = 0$ for binary networks and in our work we set $\eta = 0.05\times max(\left|W_l\right|)$ for ternary networks as in \cite{DBLP:journals/corr/ZhuHMD16}. The issue with discretization of the weights, is that it leads to the vanishing gradients problem ~\cite{DBLP:journals/corr/BengioLC13}. To overcome this, an STE is defined to replace the zero derivatives from the piecewise constant function in (\ref{weight_quant}), by a non-zero surrogate derivative~\cite{DBLP:journals/corr/HubaraCSEB16}. During training $\textbf{\textit{Q}}_l$ is used for inference and backpropagation, and the corresponding elements in $\textbf{\textit{W}}_{l}$ are updated based on these gradients. Hence the STE is defined as:
\vspace{-5pt}
\begin{align}
& \frac{\partial \hat{E}}{\partial \textit{W}_{l_{i,j}}}=\frac{\partial \hat{E}}{\partial \textit{Q}_{l_{i,j}}}
\label{STE_w}
\end{align} 
where $\hat{E}$ is the error function for a network without scaling coefficients. After training, the full precision weights are discarded and we require only the quantized weights for deployment. Whilst these methods greatly reduce computational complexity by eliminating floating point MACs, they increase the difficulty of learning. 

\subsection{Scaling}
The introduction of scaling coefficients improves learning capabilities by providing greater model capacity and compensating for the large information loss due to binary/ternary quantization. Scaling discrete weight representations requires multiplying all $\textit{Q}_{l_{i,j}}$ by positive scaling coefficients $\alpha\in\mathbb{R^+}$. We want to find optimal scaling coefficients for each layer, $\alpha_l$, which minimize our error function:
\begin{align}
& \alpha_l^* = \argminB_{\alpha} E(\alpha, \textbf{\textit{Q}})\quad s.t.\quad \alpha\geq 0,\ \textit{Q}_{l_{i,j}}\in\mathbb{C}
\end{align}
with $E$ representing the error function with scaling coefficients. Finding the optimal $\alpha_l$ is vital to reducing gradient mismatches in the forward and backward functions. It was proposed in \cite{DBLP:journals/corr/ZhouNZWWZ16} as the mean of absolute weight values for each layer: 
\vspace{-10pt}
\begin{align}
&\alpha_l = \frac{\lVert W_{l} \rVert_{1}} {Z_l}
\label{scalar_abs}
\end{align}
where $Z_l$ is the total number of layer weights. The codebook for each layer after scaling in (\ref{scalar_abs}) is symmetric: $\mathbb{\hat{C}}_l=\big\{-\alpha_l,+\alpha_l\big\}$ and the scalars become per-layer learning rate multipliers. Additionally, the STE in (\ref{STE_scalar}) reduces the gradient mismatch from (\ref{STE_w}) by including information from the full precision weights:
\vspace{-5pt}
\begin{align}
& \frac{\partial E}{\partial \textit{W}_{l_{i,j}}} = \frac{\partial E}{\partial \textit{Q}_{l_{i,j}}} = \alpha_l \frac{\partial \hat{E}}{\partial \textit{Q}_{l_{i,j}}}
\label{STE_scalar}
\end{align}
Gradient-based optimizations for scaling coefficients were also introduced in \cite{DBLP:journals/corr/ZhuHMD16} which applied different scaling coefficients for positive and negative $\textit{Q}_{l_{i,j}}$ to improve model capacity and accuracies. These are updated during backpropagation using gradients:
\vspace{-5pt}
\begin{align}
& \frac{\partial E}{\partial \alpha_{l}^p}=\sum_{i,j\in S_{l}^p}\frac{\partial E}{\partial \textit{W}_{l_{i,j}}}, \ \frac{\partial E}{\partial \alpha_{l}^n}
=  \sum_{i,j\in S_{l}^n}\frac{\partial E}{\partial \textit{W}_{l_{i,j}}}
\label{TTQ_grad}
\end{align}
where initially $\alpha_{l_0}^{p}, \alpha_{l_0}^{n} = 1$ and $S_l$ is the codebook indices for each layer, i.e. $S_{l}^p=\big\{i,j| \textit{W}_{l_{i,j}} \geq \eta  \big\}$ and $S_{l}^n=\big\{i,j| \textit{W}_{l_{i,j}} \leq -\eta  \big\}$. This allows each layer's codebook values to be asymmetric around zero, such that $\mathbb{\hat{C}}_l=\big\{-\alpha_{l}^n,+\alpha_{l}^p\big\}$. The codebook indices are then highly irregular and unordered which increases computational complexity as the matrices cannot be easily decomposed. Rather we have to check the sign of every element before computation, leading to extra branching instructions for conventional computing platforms such as CPUs/GPUs and additional logic for custom hardware. The difficulty of designing low-precision networks which have both high learning capabilities and computational efficiency can be solved by learning a symmetric codebook during training and exploiting structured matrix representations.

\section{SYQ Structural Representations}
We now propose matrix representations of SYQ by partitioning the quantization into weight subgroups. Diagonal matrix representations consist of mainly zeros and have non-zero entries along the main diagonal. For a matrix $\textbf{\textit{D}}$ to be diagonal, $\textbf{\textit{D}} = 0$ if $D_{i,j} = 0$ $\forall$ $i\neq j$, and square if $\textbf{\textit{D}}\in\mathbb{R}^{m\times m}$.  A square diagonal matrix consisting of all equal main diagonal entries is a scalar matrix. A diagonal matrix  $\boldsymbol{\alpha_l}$ is defined by the vector $\boldsymbol{\alpha_l} = \bigl[\alpha_{l}^1, ...,\alpha_{l}^{m}\bigl]$:
\[
\textbf{\textit{$\boldsymbol{\alpha}$}} = diag(\textit{$\boldsymbol{\alpha}$}):=
\begin{bmatrix}
\alpha^1 & 0 &  ..  & 0 & 0\\
0 & \alpha^2 &  ..  & : & 0\\
: & :  &  ..  & \alpha^{m-1} & :\\
0 & 0 &  ..  & 0 & \alpha^m
\end{bmatrix}
\]
Diagonal matrix multiplication is very computationally efficient as it can be easily decomposed and only the scalar vector requires storage. 
\begin{figure*}
	\begin{center}
		\includegraphics[width=3.4in]{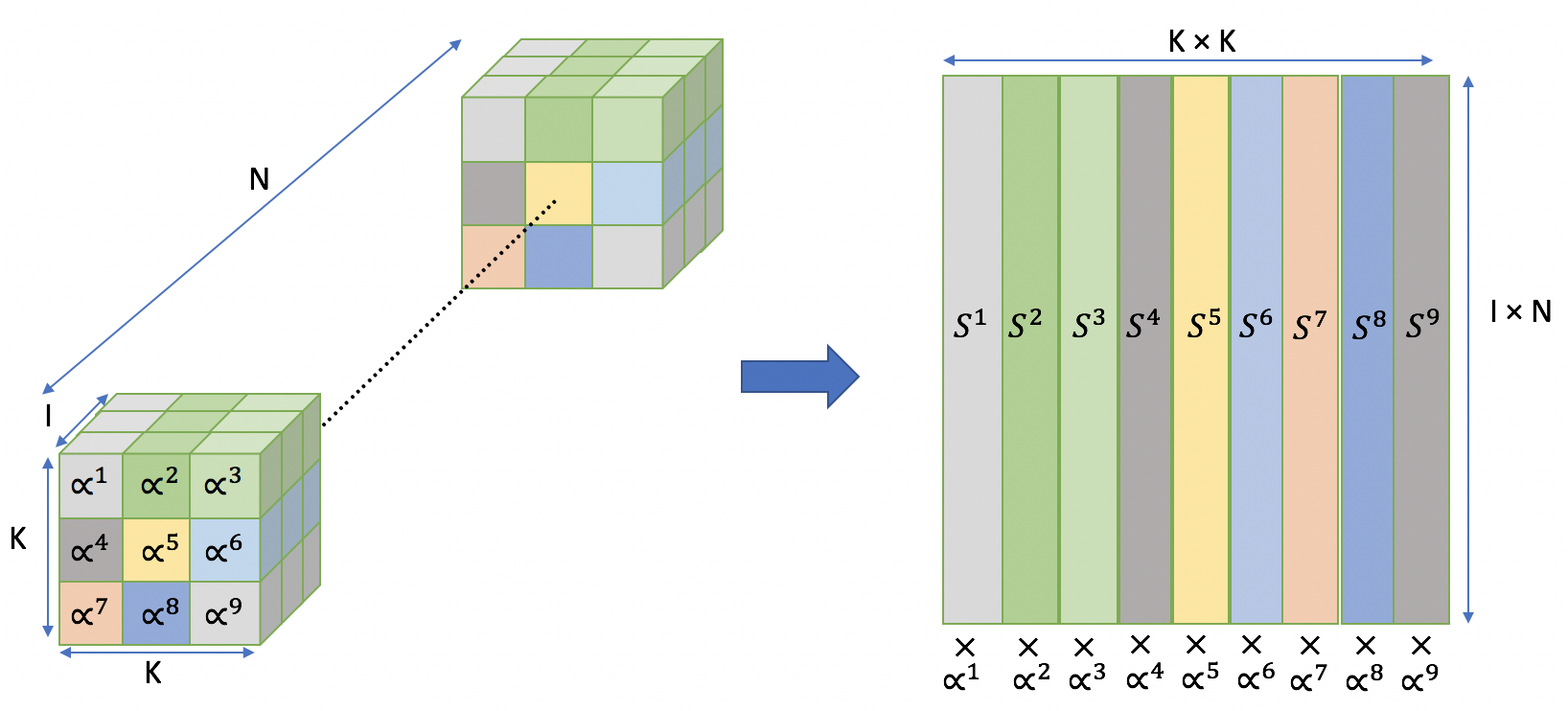}
		\includegraphics[width=3.2in]{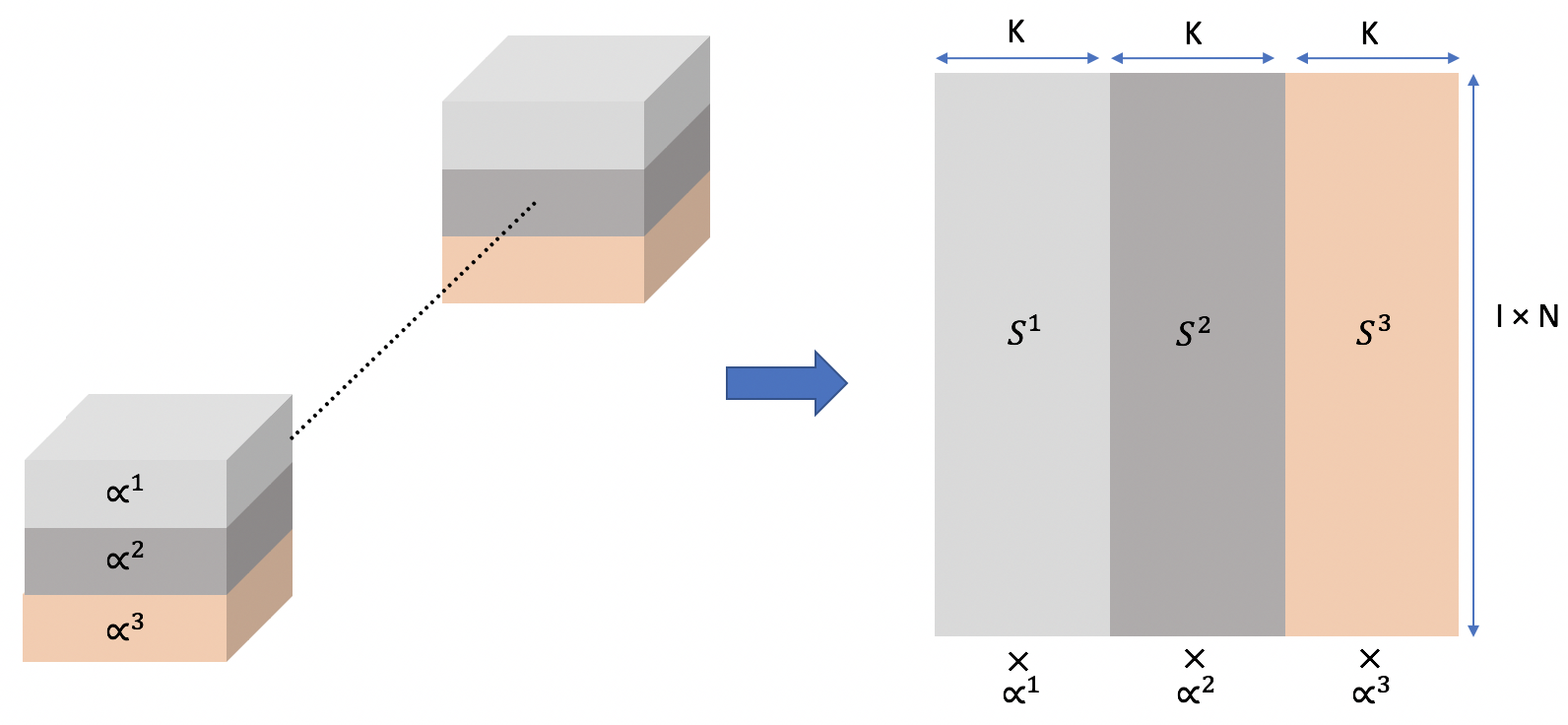}
	\end{center}
\vspace{-10pt}
	\caption{Computational structure of pixel-wise (Left) and row-wise (Right) subgrouping of a CONV layer (K, I = 3). The tensors represent the weight layer structure during training and the matrices represent the matrix decomposition for deployment.}
	\label{fig:short}
	\vspace{-5pt}
\end{figure*} 

\subsection{Layers}
CONV and FC layers have differing computational requirements and sensitivities to network redundancies. CONV weights are reused many times across the input feature map whereas FC weights are used only once per image. Hence, the quantization error of each weight in a CONV layer impacts the dot products across the entire input feature map volume rather than just once for FC weights. Thus, a fine-grained approach to CONV layers is effective at compensating for this error. Quantized CONV weights are represented as a tensor $\textbf{Q}_{l}\in\mathbb{R}^{Z}$ with $Z=K\times K\times I\times N$.  As typically $I,N \gg K$, it is optimal to have a diagonal scalar of size $K\times K$ or even $K^2\times K^2$ as only small scalar vectors are required for storage. By reshaping the tensor $\textbf{Q}_{l}$, we form a matrix $\textbf{\textit{Q}}_{l}\in\mathbb{R}^{\hat{Z}}$ where $\hat{Z}= K^2\times(I N)$ or $\hat{Z}= K\times(INK)$ and represent our scalar matrix multiplication as $diag(\boldsymbol{\alpha_l})\textbf{\textit{Q}}_l^T$ with the square diagonal matrix, $ diag(\boldsymbol{\alpha_l})\in\mathbb{R}^{K^2\times K^2}$ or $ diag(\boldsymbol{\alpha_l})\in\mathbb{R}^{K\times K}$ respectively. FC layers are represented as a matrix $\textbf{\textit{Q}}_{l}\in\mathbb{R}^{L\times H}$ where $H$ is the number of hidden nodes and $L$ the activation neurons.  As FC layers are more robust to quantization, one learnable scaling coefficient (layer-wise) for the FC layer can sufficiently approximate the distribution and also can be represented with scalar matrix computation. All elements in $\boldsymbol{\alpha_l}$ are then equal and we only require storage of one value.

\subsection{Subgroups}
More fine-grained quantization can improve approximations of the statistical distributions of weights. We implement pixel-wise scaling for CONV layers which involves grouping all spatially equivalent pixels along the $I\times N$ dimension. This results in different values for all the main diagonal elements in $ diag(\boldsymbol{\alpha})\in\mathbb{R}^{K^2\times K^2}$. With this representation, we can still decompose the matrix computation along each pixel dimension and exploit the parallel nature of convolutions as shown in Figure~\ref{fig:short}. We do this by creating subgroups $1\leq i \leq K^2$ with codebook indices $S_l^i=\big\{j| \textit{W}_{l_{i,j}}\big\}$. Other granularities such as row-wise scaling involve grouping all pixels along a row or column ($I\times N\times K$), resulting in $S_l^j = S_{l}^i\cup S_{l}^{i+1}...\cup S_{l}^{K}$ where $1\leq j \leq K$ (as illustrated in Figure~\ref{fig:short}) and also layer-wise scaling: $S_l = S_{l}^i\cup S_{l}^{i+1}...\cup S_{l}^{K^2}$. Different granularities affect both accuracy and computation as further explored in Sections 6 $\&$ 7.

\section{SYQ Training}  
In this section, we now describe the methodology to efficiently train SYQ networks.

\subsection{Symmetric Quantizer}
When training low-precision inference networks, the aim is to have the smallest possible codebook. Typically, as the codebook size increases, a network will approach full-precision performance but increase hardware cost. However, there are certain codebook representations which are significantly more hardware friendly than others and won't necessarily impose any hardware costs. Given a codebook $C$, and the nonzero codebooks $C^p=\big\{c_i| c_i > 0 \big\}$ and $C^n=\big\{c_j| c_j < 0 \big\}$, a quantizer is denoted as symmetric if:
\begin{align}
& \forall c_i\in C^p, \quad \exists \left|c_j\right|\in C^n\quad \text{where}\quad c_i=\left|c_j\right|
\end{align}
Learning this type of codebook requires updating one scaling coefficient during training for two bi-polar codebook values. The gradient of each scaling coefficient for each subgroup becomes:
\vspace{-5pt}
\begin{align}
& \frac{\partial E}{\partial \alpha_{l}^i}=\sum_{j\in S_{l}^i}\frac{\partial E}{\partial \textit{W}_{l_{i,j}}}
\label{SYQ_grad}
\end{align}
When computing binary/ternary weight representations followed by a scale, it is ideal to have a codebook which is symmetric around zero, as the codebook storage requirements are almost halved. This is because only the absolute value of the two symmetric values needs to be stored. Additionally, codebook indices become highly regular and ordered for the scalar multiply which greatly reduces computational complexity. The nature of symmetric quantization enables the opportunity to implement fine-grained quantization (pixel/row-wise) whilst maintaining the scalar matrix multiplication structure used in layer-wise scaling. This is also advantageous as the scaling coefficients become fine-grained adaptive learning rate multipliers for each pixel/row in a CONV layer, i.e. the STE becomes:
\vspace{-2pt}
\begin{align}
& \frac{\partial E}{\partial \textit{W}_{l_{i,j}}} = \frac{\partial E}{\partial \textit{Q}_{l_{i,j}}}
=  \alpha_l^i \frac{\partial \hat{E}}{\partial \textit{Q}_{l_{i,j}}}
\label{lr_multipliers}
\end{align} 
As the use of scaling coefficients can more accurately approximate subgroups and are gradient-based, the gradient mismatch is significantly reduced for weight quantization which enhances network learning.

\subsection{Initialization}
The solution to non-convex optimizations such as gradient descent depend heavily on parameter initialization to avoid vanishing or exploding activations/gradients and ensure network convergence~\cite{Glorot10understandingthe}. For low-precision networks, excessive gradient mismatches between the forward and backward functions must be minimized, otherwise the gradients will not propagate well. To deal with this concern, the scaling coefficients coefficients are initialized as the mean of full precision weights in it's corresponding subgroup. For example, the scaling coefficient in pixel-wise scaling is: 
\vspace{-7pt}
\begin{align}
&\alpha_{l_0}^i =\frac{\sum_{j\in S_{l}^i}\left|W_{l_{i,j}}\right|}{I\times N}
\label{eq:initial}
\end{align} 
Layer-wise scaling in FC layers has $\alpha_{l_0}$ as the mean of all layer weights. By incorporating information from the full precision weights, we aim to reduce the mismatch initially and the scaling coefficients are then optimized during backpropagation.

\subsection{Activations Quantization}
Our forward path approximation to $g$ in (\ref{eq:DNN_act}) uniformly quantizes a real number $x\in[0,M]$ to a k-bit number:
\begin{align}
& G(x) = \frac{1}{2^f}floor((2^f)x+\frac{1}{2})
\label{eq:actquant}
\end{align}
where $floor$ represents the round down operation and $M$ is the upper bound. $M$ itself is bounded by its arbitrary unsigned two's complement fixed point representation where $f$ is the number of fractional bits and $M=2^{k-f} - 2^{-f}$. Uniform quantization translates to a reduction in hardware implementation complexity. To achieve this, we use the following STE for the activations:
\begin{align}
& \frac{\partial E}{\partial x}=\frac{\partial E}{\partial G}
\label{act_grad}
\end{align} 
Differences in the forward and backward activation functions create a gradient mismatch which can result in unstable and inefficient learning. To minimize this issue, we adjust $M$ as a hyperparameter. The overall SYQ training process is summarized in Algorithm~\ref{alg:one}. 
\begin{algorithm}[tb]
	\caption{SYQ Training Summary For DNNs.}
	\label{alg:one}
	\begin{algorithmic}
		\STATE {\bfseries Initialize:} Set subgrouping granularity for $S_{l}^i$  and set $\alpha_{l_0}^i$.\STATE {\bfseries Inputs:} Minibatch of inputs \& targets $(I,Y)$, Error function $E(Y,\hat{Y})$, current weights $\boldsymbol{W_t}$ and learning rate, $\gamma_t$\STATE {\bfseries Outputs:} Updated $\boldsymbol{W_{t+1}}$, $\boldsymbol{\alpha_{t+1}}$ and $\gamma_{t+1}$ 
		\STATE
		\STATE {\bfseries $\textit{SYQ Forward:}$}
		\FOR{l=1 to L}
		\STATE $\textbf{\textit{Q}}_l=sign(\textbf{\textit{W}}_l)\odot\textbf{\textit{M}}_l$ \textbf{with} $\eta$, using (\ref{weight_quant}) \& (\ref{mask})\\
		\FOR{ith subgroup in lth layer}
		\STATE Apply $\alpha_{l}^i$ to $S_{l}^i$
		\ENDFOR
		\ENDFOR
		\STATE $\hat{Y}$ = {\bfseries SYQForward} ($ I, Y, \textbf{\textit{Q}}_l, \boldsymbol{\alpha}_l$) using (\ref{eq:actquant})
		\STATE 
		\STATE {\bfseries $\textit{SYQ Backward:}$}
		\STATE $\frac{\partial \hat{E}}{\partial \textbf{\textit{Q}}_l}$ = {\bfseries WeightBackward}($\textbf{\textit{Q}}_l,\boldsymbol{\alpha}_l,\frac{\partial \hat{E}}{\partial \hat{Y}})$ using (\ref{lr_multipliers}) \& (\ref{act_grad})
		\STATE $\frac{\partial \hat{E}}{\partial \boldsymbol{\alpha}_l}$ = {\bfseries ScalarBackward}($\frac{\partial \hat{E}}{\partial \textbf{\textit{Q}}_l},\boldsymbol{\alpha}_l,\frac{\partial \hat{E}}{\partial \hat{Y}})$ using (\ref{SYQ_grad})
		\STATE $\boldsymbol{W_{t+1}}$ = {\bfseries UpdateWeights}($\boldsymbol{W_t}, \frac{\partial \hat{E}}{\partial \textbf{\textit{Q}}_l}, \gamma$)
		\STATE $\boldsymbol{\alpha_{t+1}}$ = {\bfseries UpdateScalars}($\boldsymbol{\alpha_t}, \frac{\partial \hat{E}}{\partial \boldsymbol{\alpha}_l}, \gamma$)
		\STATE $\gamma_{t+1}$ = {\bfseries UpdateLearningRate}($\gamma_t, t$)
	\end{algorithmic}
\end{algorithm}

\section{Experiments}
To demonstrate the versatility of SYQ, we applied it to several state-of-the-art benchmark models, all with different network topologies. We use binary/ternary weights and varying activation bitwidths for classification of the large-scale ImageNet dataset. The ILSVRC-2012 ImageNet is a natural high resolution visual classification dataset consisting of 1000 classes, 1.28 million training images and 50K validation images. Inputs are resized to $256\times 256$ before being randomly cropped to $224\times 224$. We report our single-crop evaluation results using Top-1 and Top-5 accuracy.

\subsection{Networks}
We compare our results to the full precision baseline and benchmark reference model accuracies in Table \ref{networks}\footnote{Our ResNet and AlexNet reference results are obtained from https://github.com/facebook/fb.resnet.torch and https://github.com/BVLC/caffe, respectively}, showing that SYQ training achieves similar accuracy to floating point. This suggests the noise induced from replacing floating point weight layers with SYQ versions, provides effective regularization during training. An AlexNet~\cite{NIPS2012_4824} variant is implemented which eliminates dropout and includes batch normalization~\cite{DBLP:journals/corr/IoffeS15}. A mini batch size of 64 is used, L2 weight decay of 5e-6, and our learning rate is initially 1e-4 with step decays of scale factor 0.2. For ResNet \cite{DBLP:journals/corr/HeZRS15}, we test on the 18, 34 and 50 layer variations. Our batch size is 128, learning rate is initially 1e-3 with step decay of factor 0.2. We also test on a variant of VGG-16 \cite{DBLP:journals/corr/SimonyanZ14a}, using model-A in \cite{DBLP:journals/corr/HeZR015} with the spp layer replaced by a max pool and only 3 CONV layers rather than 5 for input size blocks of 56, 28 and 14, as in \cite{DBLP:journals/corr/CaiHSV17}. Batch sizes are set to 32 and our learning rate is initially 1e-4 with a step decay of factor 0.2. The VGG and ResNet models were initialized from floating point baseline weights. Full-precision weights are used for the first and last layer. All other CONV layers are quantized with SYQ pixel-wise scaling, FC layers with layer-wise scaling and the activations of all layers using (\ref{eq:actquant}).
\begin{table}[t]
	\caption{Summary of Results for 8-bit activations and binary (1-8) and ternary (2-8) weights}
	\vspace{-15pt}
	\label{networks}
	\begin{center}
		\begin{tabular}{l|l|l|l|l|l}  
			\multicolumn{1}{c}{\bf Model}  
			&&\multicolumn{1}{c}{\bf 1-8}
			&\multicolumn{1}{|c}{\bf 2-8}
			&\multicolumn{1}{|c}{\bf Baseline}
			&\multicolumn{1}{|c}{\bf Reference}\\ \hline
			\multirow{2}{*}{AlexNet} &Top-1&\bf{56.6}&\bf{58.1}& 56.6&57.1 \\
			&Top-5&\bf{79.4}&\bf{80.8}&80.2&80.2\\
			\hline
			\multirow{2}{*}{VGG} &Top-1&\bf{66.2}&\bf{68.7}& 69.4&- \\
			&Top-5&\bf{87.0}&\bf{88.5}&89.1&-\\
			\hline
			\multirow{2}{*}{ResNet-18} &Top-1&\bf{62.9}&\bf{67.7}& 69.1&69.6 \\
			&Top-5&\bf{84.6}&\bf{87.8}&89.0&89.2\\
			\hline
			\multirow{2}{*}{ResNet-34} &Top-1&\bf{67.0}&\bf{70.8}& 71.3&73.3 \\
			&Top-5&\bf{87.6}&\bf{89.8}&89.1&91.3\\
			\hline
			\multirow{2}{*}{ResNet-50} &Top-1&\bf{70.6}&\bf{72.3}& 76.0 &76.0 \\
			&Top-5&\bf{89.6}&\bf{90.9}&93.0&93.0\\
			\hline
		\end{tabular}
	\end{center}
\vspace{-20pt}
\end{table}

\subsection{Changing Granularity Via Weight Subgroups}
Weight subgroups can be arbitrarily designed for a given hardware application. Table \ref{granularity} shows accuracy differences between using row/layer-wise vs pixel-wise scaling on AlexNet and suggests pixel-wise and row-wise are marginally different, especially for higher precisions, but both are considerably more accurate than layer-wise. 
\begin{table}[t]
	\caption{AlexNet accuracy differences between using row/layer-wise and pixel-wise symmetric quantization}
	\vspace{-5pt}
	\label{granularity}
	\begin{center}
		\begin{tabular}{|ll|ll|ll|}
			\hline
			&& \multicolumn{2}{|c|}{Row-wise} 
			& \multicolumn{2}{|c|}{Layer-wise} \\
			\hline
			\multicolumn{1}{|c}{\bf Weights}
			&\multicolumn{1}{c}{\bf Act.}
			&\multicolumn{1}{|c}{\bf Top-1}
			&\multicolumn{1}{c}{\bf Top-5}
			&\multicolumn{1}{|c}{\bf Top-1}
			&\multicolumn{1}{c|}{\bf Top-5}
			\\ \hline
			1&2&-0.7&-0.5&-1.4&-2.2\\
			1&8&-0.1&-0.3&-0.4&-2.2\\
			2&2&+0.1&-0.0&-1.3&-1.5\\
			2&8&-0.1&-0.1&-1.9&-1.7\\
			\hline
		\end{tabular}
	\end{center}
\vspace{-20pt}
\end{table}
This demonstrates the effectiveness of fine-grained quantization of CONV layers over layer-wise and promotes the exploration for efficient representations of scalar computation. It also shows the effectiveness of row-wise quantization as it typically incurs a smaller memory requirement with a small accuracy drop, for a significant gain in the potential parallelism of the network. 

\subsection{Comparisons To Previous Work}
We compare SYQ explicitly using AlexNet, ResNet-18 and ResNet-50 in Tables~\ref{AlexNet}, \ref{Res18} \& \ref{Res50} as they've been extensively studied in the literature. Our ternary results with 8 bit activations (2w-8act) improves on the state-of-the-art for all three networks. Our 2w-4act for ResNet-50 also improves on the state-of-the-art FGQ. This is also the case for binary weights, such as 1w-8act ResNet-18 and AlexNet with 1w-2/4act. For extremely low 1w-2act representations, SYQ also has a 2.7\% increase in Top-1 accuracy over the state-of-the-art HWGQ. This demonstrates SYQ's superiority for producing high accuracy. Additionally, it shows that multiple learnable scaling coefficients effectively reduce the gradient mismatch in the forward and backward paths, translating to efficient learning under low-precision constraints. 
\begin{table}[t]
	\caption{Comparison to previously published AlexNet results}
	\vspace{-5pt}
	\label{AlexNet}
	\begin{center}
		\begin{tabular}{lllll}
			\multicolumn{1}{c}{\bf Model}  &\multicolumn{1}{c}{\bf Weights}
			&\multicolumn{1}{c}{\bf Act.}
			&\multicolumn{1}{c}{\bf Top-1}
			&\multicolumn{1}{c}{\bf Top-5}
			\\
			\hline
			DoReFa-Net~\cite{DBLP:journals/corr/ZhouNZWWZ16}&1&2&49.8&- \\
			QNN~\cite{DBLP:journals/corr/HubaraCSEB16} &1&2&51.0&73.7 \\
			HWGQ~\cite{DBLP:journals/corr/CaiHSV17} &1&2&52.7&76.3 \\
			\hline
			\bfseries{SYQ} &\bfseries{1} &\bfseries{2}&\bfseries{55.4} & \bfseries{78.6}\\
			\hline
			DoReFa-Net~\cite{DBLP:journals/corr/ZhouNZWWZ16}&1&4&53.0&- \\
			\hline
			\bfseries{SYQ} &\bfseries{1} &\bfseries{4} &\bfseries{56.2} & \bfseries{79.4}\\
			\hline
			BWN~\cite{DBLP:journals/corr/RastegariORF16}&1 &32&56.8&79.4\\
			\hline
			\bfseries{SYQ} &\bfseries{1} &\bfseries{8} &\bfseries{56.6} & \bfseries{79.4}\\
			\hline
			\bfseries{SYQ} &\bfseries{2} &\bfseries{2} &\bfseries{55.8} & \bfseries{79.2}\\
			\hline
			FGQ~\cite{DBLP:journals/corr/MellempudiKM0KD17}&2 &8&49.04&-\\
			TTQ~\cite{DBLP:journals/corr/ZhuHMD16}&2 &32&57.5&79.7\\
			\hline
			\bfseries{SYQ} &\bfseries{2} &\bfseries{8} &\bfseries{58.1} & \bfseries{80.8}\\
			\hline
		\end{tabular}
	\end{center}
\vspace{-15pt}
\end{table}

\begin{table}[t]
	\caption{Comparison to previously published ResNet-18 results}
	\vspace{-5pt}
	\label{Res18}
	\begin{center}
		\begin{tabular}{lllll}
			\multicolumn{1}{c}{\bf Model}  &\multicolumn{1}{c}{\bf Weights}
			&\multicolumn{1}{c}{\bf Act.}
			&\multicolumn{1}{c}{\bf Top-1}
			&\multicolumn{1}{c}{\bf Top-5}
			\\ \hline
			BWN~\cite{DBLP:journals/corr/RastegariORF16}  &1 &32&60.8&83.0\\
			\hline
			\bfseries{SYQ} &\bfseries{1} &\bfseries{8} &\bfseries{62.9} & \bfseries{84.6}\\
			\hline
			TWN~\cite{DBLP:journals/corr/LiL16} &2&32&65.3&86.2 \\
			INQ~\cite{DBLP:journals/corr/ZhouYGXC17} &2&32&66.0&87.1 \\
			TTQ~\cite{DBLP:journals/corr/ZhuHMD16}&2 &32&66.6&87.2\\
			\hline
			\bfseries{SYQ} &\bfseries{2} &\bfseries{8} &\bfseries{67.7} & \bfseries{87.8}\\
			\hline
		\end{tabular}
	\end{center}
\vspace{-18pt}
\end{table}
\begin{table}[t]
	\caption{Comparison to previously published ResNet-50 results}
	\vspace{-5pt}
	\label{Res50}
	\begin{center}
		\begin{tabular}{lllll}
			\multicolumn{1}{c}{\bf Model}  &\multicolumn{1}{c}{\bf Weights}
			&\multicolumn{1}{c}{\bf Act.}
			&\multicolumn{1}{c}{\bf Top-1}
			&\multicolumn{1}{c}{\bf Top-5}
			\\ \hline
			HWGQ~\cite{DBLP:journals/corr/CaiHSV17} &1&2&64.6&85.9 \\
			\hline
			\bfseries{SYQ} &\bfseries{1} &\bfseries{4} &\bfseries{68.8} &
			\bfseries{88.7}\\
			\hline
			\bfseries{SYQ} &\bfseries{1} &\bfseries{8} &\bfseries{70.6} & \bfseries{89.6}\\
			\hline
			FGQ~\cite{DBLP:journals/corr/MellempudiKM0KD17} &2&4&68.4&- \\
			\hline
			\bfseries{SYQ} &\bfseries{2} &\bfseries{4} &\bfseries{70.9} & \bfseries{90.2}\\
			\hline
			FGQ~\cite{DBLP:journals/corr/MellempudiKM0KD17} &2&8&70.8&- \\
			\hline
			\bfseries{SYQ} &\bfseries{2} &\bfseries{8} &\bfseries{72.3} & \bfseries{90.9}\\
			\hline
		\end{tabular}
	\end{center}
\vspace{-20pt}
\end{table}

\subsection{Varying Activation Bitwidth}
\begin{figure}
	\begin{center}
		\includegraphics[width=2.8in]{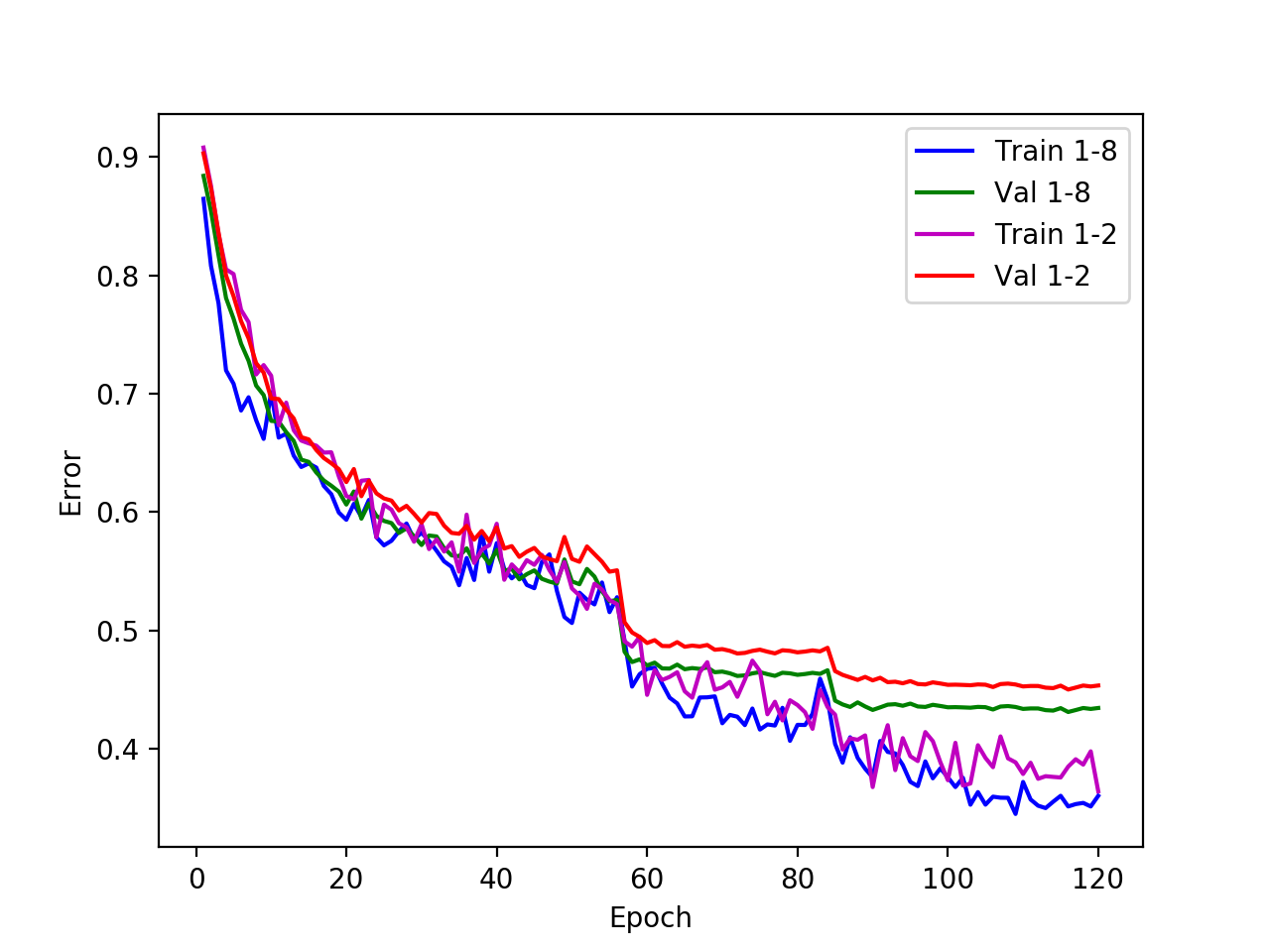}
	\end{center}
	\vspace{-10pt}
	\caption{Top-1 training and validation error for binary AlexNet with varying activation precisions}
	\label{fig:training}
	\vspace{-5pt}
\end{figure}

The most important result is that SYQ efficiently quantizes networks with low-precisions for both weights and activations. From Figure~\ref{fig:training},
we can see that lowering the precision of the activations does not severely alter the training curve, suggesting that the gradient information from pixel-wise scaling coefficients in SYQ compensates well for the loss of information. However, when quantizing down to 2-bits, the training error curve does become more volatile, demonstrating instabilities in network learning. We also report the classification accuracies for varying activations and bitwidths on AlexNet and ResNet-50 in Tables~\ref{AlexNet} \&~\ref{Res50}, which shows that there is minimal discrepancy from the full-precision networks with as low as 4-bit activations. These results are extremely promising and have strong implications for specialized hardware implementations of low-power DNNs.

\section {Hardware Implications}
In this section we discuss the computational implications of different scaling operations and present a design for specialized hardware implementations.

\subsection{Computational and Memory Complexity}
Considering a CONV layer with Ops, $P = K\times K\times I\times N\times F\times F$, where $F$ is the IFM dimension. The layer-wise scaling, as in DoReFa-Net, requires one scaling coefficient per $P$ operations. 
\begin{table}[t]
	\caption{Number of scaling coefficients and operations per layer, for different techniques}
	\vspace{-5pt}
	\label{complexity}
	\begin{center}
		\begin{tabular}{l|ll}
			\hline
			\multicolumn{1}{c|}{\bf Method}
			&\multicolumn{1}{c}{\bf Scalars}
			&\multicolumn{1}{c}{\bf Ops}
			\\ \hline
			Layer (DoReFa)&1&$P$\\
			Row (SYQ)&$K$&$P$\\
			Pixel (SYQ)&$K^2$&$P$\\
			Asymmetric (TTQ)&2&$P+Z$\\
			Grouping (FGQ)&$K^2N/4$&$P$\\
			Channel (HWGQ/BWN)&$N$&$P$\\
			\hline
		\end{tabular}
	\end{center}
\vspace{-25pt}
\end{table}
For channel-wise scaling in HWGQ and BWN, it requires N scaling coefficients as there is one per output feature map, where typically $N \gg 1$. TTQ implements asymmetric layer-wise quantization which requires two scaling coefficients per layer and $P+Z$ operations as we add a branching operation for each weight due to irregular codebook indices, as described in Section 3.3. FGQ uses pixel-wise scaling for every 4 filters, whereas SYQ uses pixel-wise scaling per $N$ filters, hence it requires $K^2N/4$ scaling coefficients and $P$ operations. For pixel-wise SYQ scaling, $K^2$ scaling coefficients and $P$ operations are required, where $K=3$ for most CONV layers in modern networks. For row-wise SYQ scaling it requires $K$ scaling coefficients and $P$ operations. These results are displayed in Table \ref{complexity}, demonstrating the benefits of maintaining a diagonal representation for the scalar matrix multiplication of each layer as we either improve computational or memory complexity against all other fine-grained methods. Another key benefit of SYQ is its amenability to highly parallel processors.

\subsection{Architectural Design}
\begin{figure}[t]
	\begin{center}
		\includegraphics[width=3.0in]{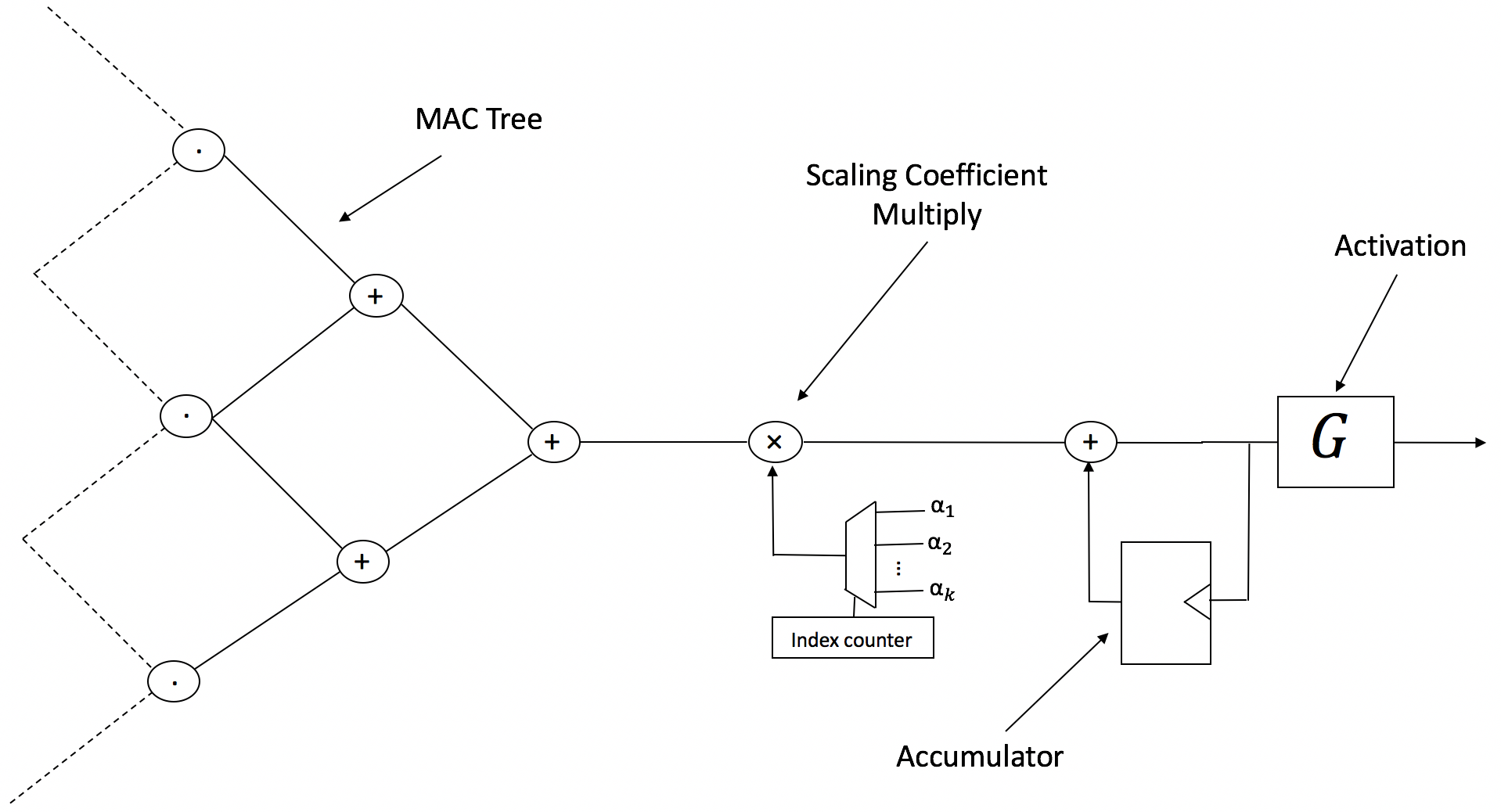}
	\end{center}
\vspace{-12pt}
	\caption{Hardware description of MAC for SYQ layers}
	\label{fig:hardware}
	\vspace{-5pt}
\end{figure}

For the CONV layer, the operations are a sum of dot products between the input and kernel filter. In order to reduce compute complexity, we increase the number of operations in each dot product, while significantly decreasing the complexity of each operation.
For example, the size of the input vector, in the calculation of each dot product is: $L_v = K^2 I$.
The number of operations is $Op_{mul}^{L} = L_v$ for multiplies and $Op_{add}^{L} = L_v-1$ for additions. Given that we have a limited codebook for our weights, we can break it into sub-dot products where we apply the scaling factor, $\alpha^i$, after we have computed the sub-dot product for that set of symmetrically constrained weights.
For pixel-wise quantization, the total multiplies becomes $Op_{mul}^{P} = L_v + K^2$ and the total adds become $Op_{add}^{P} = K^2(L_v / K^2 - 1) + (K^2 - 1) = L_v - 1$.
However, the first term in each of these calculations can be done at significantly lower precision. For multiplies this means a binary or ternary multiple - which can often be implemented as a bit-flip.
To compute this in specialized hardware, for layer-wise scaling, we have a parallel MAC tree which consists of a multiply of an input and binary/ternary number (represented as a dot) followed by an adder tree to sum up the outputs.
Outputs of these are fed into a multiplier to compute the scale, followed by an accumulator to store the outputs before being fed into the activation function. This architecture is shown in Figure~\ref{fig:hardware}.
For every hardware block of this type, our per-pixel/row scaling only requires one additional ring counter which stores scaling coefficients and shifts the input to the scaling multiplier through an index counter as each row/pixel is finished computing which is computationally inexpensive.
As in the equivalent layer-wise scaling architecture, we can still maintain one multiplier in hardware and only increase memory slightly to store the scaling coefficients.
Table~\ref{tab:fpga_impl} shows the resource and performance estimates provided by Vivado HLS of the described hardware architecture for a target Xilinx ZU3 FPGA device at an estimated clock frequency of over 300 MHz.
The main design is based on the MVTU described in FINN~\cite{DBLP:journals/corr/UmurogluFGBLJV16}, with an extension to 2-bit activations and pixel-wise and row-wise SYQ.
The layer-wise baseline uses no multiplies, as these can absorb into quantization thresholds for activations~\cite{DBLP:journals/corr/UmurogluFGBLJV16}.
The MVTU was configured for a convolution layer with $I=384$, $N=256$, $K=3$, while scaling the size of the MAC tree (SIMD) and the number of parallel processors (PE).
As shown, the BRAM (memory blocks on an FPGA (18k)) and LUT usage is almost identical, while the DSP usage increases proportionally with the number of parallel output channels which are processed.  The increase in DSPs is not necessarily costly for the ZU3 as we are able to utilize more of the total available resources. Resource usage is only shown for pixel-wise SYQ, as row-wise only differed in LUT usage by less than 2\%.

\begin{table}[t]
	\caption{Resource Usage of a Matrix-Vector Processing Unit with Layer-wise and Pixel-wise Quantization for target Xilinx ZU3}
	\label{tab:fpga_impl}
	\begin{center}
		\begin{tabular}{l|llllll}
			\hline
            Config & SIMD      & PE        & BRAMs & LUTs (k) & DSPs\\
			\hline
            Layer  & 32        & 32        & 64        & 29.8 & 4   \\
            Layer  & 64        & 32        & 64        & 56.5 & 4   \\
            Layer  & 32        & 64        & 64        & 58.9 & 4   \\
            SYQ(P) & 32        & 32        & 64        & 29.4 & 36  \\
            SYQ(P) & 64        & 32        & 64        & 56.1 & 36  \\
            SYQ(P) & 32        & 64        & 64        & 57.7 & 68  \\
            \hline
            ZU3    & -         & -         & 432       & 70.6 & 360 \\
			\hline
		\end{tabular}
	\end{center}
\vspace{-20pt}
\end{table}
\section{Conclusions}
The problem of efficiently training large DNNs with low-precision weights and activations is considered. We propose learning symmetric quantization for DNNs in order to maximize network learning whilst minimizing hardware complexity. This was achieved by constraining the solution to low-precision representations and learning a diagonal scalar matrix using gradient-based optimizations for efficient computation. As a result, we reduce the computational requirements of fine-grained quantization and achieve state-of-the-art accuracies on modern benchmark networks.

\section*{Acknowledgements}
This research was partly supported under the Australian Research
Councils Linkage Projects funding scheme (project number LP130101034) and Zomojo
Pty Ltd.
{\small
\bibliographystyle{ieee}
\bibliography{egbib}
}

\end{document}